\documentclass[a4paper, 10pt, conference]{ieeeconf}      

\IEEEoverridecommandlockouts                              

\usepackage[T1]{fontenc}
\usepackage[utf8x]{inputenc}
\usepackage{amssymb}
\usepackage{xcolor}
\usepackage{comment}

\usepackage{url}
\usepackage{booktabs}
\usepackage{bbding}
\usepackage{pifont}
\usepackage{utfsym}
\usepackage{fontawesome}
\usepackage{multirow}
\usepackage{arydshln}
\usepackage{amsmath} 
\usepackage{soul}
\usepackage{hyperref}

\title{\LARGE \bf
A Dataset and Benchmarks for Deep Learning-Based Optical Microrobot Pose and Depth Perception
}

\author{Lan Wei and Dandan Zhang
\thanks{*Additional dataset details are available on our website: \textit{https://lannwei.github.io/Optical-Microrobot-Database/}}
\thanks{
Lan Wei and Dandan Zhang are with the Imperial-X Initiative, Department of Bioengineering, Imperial College London, London, United Kingdom.
Corresponding: Dandan Zhang, {\tt\small d.zhang17@imperial.ac.uk}}
}

\begin{document}

\maketitle
\thispagestyle{empty}
\pagestyle{empty}

\begin{abstract}

Optical microrobots, manipulated via optical tweezers (OT), have broad applications in biomedicine. However, reliable pose and depth perception remain fundamental challenges due to the transparent or low-contrast nature of the microrobots, as well as the noisy and dynamic conditions of the microscale environments in which they operate. An open dataset is crucial for enabling reproducible research, facilitating benchmarking, and accelerating the development of perception models tailored to microscale challenges. Standardised evaluation enables consistent comparison across algorithms, ensuring objective benchmarking and facilitating reproducible research.
Here, we introduce the \underline{O}p\underline{T}ical \underline{M}icro\underline{R}obot dataset (OTMR), the first publicly available dataset designed to support microrobot perception under the optical microscope. OTMR contains 232,881 images spanning 18 microrobot types and 176 distinct poses.
We benchmarked the performance of eight deep learning models, including architectures derived via neural architecture search (NAS), on two key tasks: pose classification and depth regression. Results indicated that Vision Transformer (ViT) achieve the highest accuracy in pose classification, while depth regression benefits from deeper architectures. Additionally, increasing the size of the training dataset leads to substantial improvements across both tasks, highlighting OTMR’s potential as a foundational resource for robust and generalisable microrobot perception in complex microscale environments.

\end{abstract}

\section{Introduction}
\begin{figure}[!t]
\centering
\includegraphics[width=1\hsize]{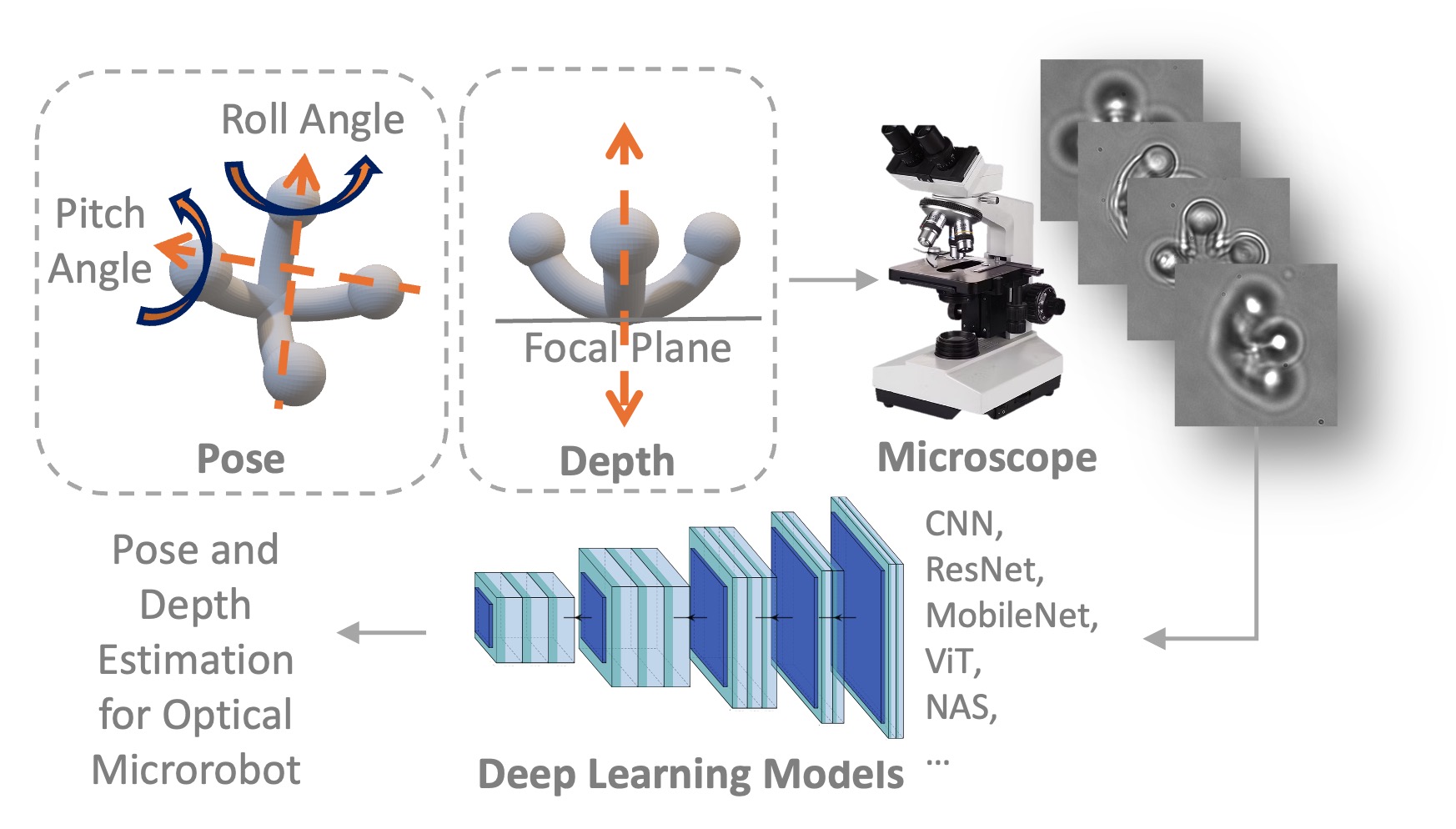}
\vspace{-0.6cm}
\caption{
Conceptual overview of out-of-plane pose and depth estimation for optical microrobots. 
The pose is defined by pitch and roll angles, while depth corresponds to the vertical displacement relative to the microscope’s focal plane. 
Images are captured using an optical microscope and used to train deep learning models. 
These models learn to estimate the microrobot’s out-of-plane pose and depth from 2D microscopy images.
}
\label{fig-concept}
\vspace{-0.5cm}
\end{figure}

Optical tweezers (OT) have been used to manipulate cells and other microscopic biological objects with high precision, contributing to advances in tissue engineering, micro-assembly, and various other areas of biomedical research~\cite{grier2003revolution,chowdhury2013automated,zhang2022fabrication}.
Despite these advantages, the use of focused laser beams can lead to thermal damage and photothermal effects, which hinder safe interaction with sensitive biological materials~\cite{blazquez2019optical, li2021opto}. 
To mitigate these issues, researchers have explored the use of optical microrobots actuated by OT for indirect manipulation, which reduces the risk of mechanical damage to biological samples and allows for precise control in dynamic microscale environments~\cite{zhang2020distributed}. 
Nonetheless, working at the microscale introduces a range of complex physical phenomena—such as nonlinear optical trapping behaviour, nonuniform fluid dynamics, Brownian motion, and surface forces like van der Waals interactions that make precise microrobot control particularly challenging~\cite{xie2019reconfigurable, kundu2019measurement}. 
Achieving effective closed-loop control systems further necessitates accurate, real-time estimation of the microrobot’s out-of-plane pose and depth (as shown in Fig.~\ref{fig-concept})~\cite{zhang2022micro, sha2019review}. 
However, common issues in microscopic imaging, such as defocusing, optical diffraction, and background noise, can affect reliable tracking and pose estimation during manipulation tasks~\cite{muinos2021reinforcement}.

Deep learning-based methods offer a promising solution for microrobot perception, as they can automatically extract hierarchical visual features and learn complex mappings from raw images to pose and depth representations~\cite{chen2023deep,choudhary2024three}. 
However, unlike perception tasks in the macroscopic world, where objects typically exhibit clear contours and distinct textures, 
microscale targets often lack
distinguishable visual features, thereby presenting substantial challenges for accurate perception.
Moreover, prior studies have primarily validated their methods on a single microrobot model, without evaluating performance across a variety of robots with diverse and intricate geometries~\cite{zhang2022micro}. 
This limits the generalisability of the proposed approaches. 
Developing robust models requires access to large-scale and diverse datasets. 
In microrobotics, however, acquiring such datasets is particularly challenging. The high costs associated with micro/nano-fabrication, combined with the technical difficulties of accurately capturing out-of-plane poses across different microrobot types, make large-scale data acquisition both expensive and labour-intensive~\cite {li2024control}. 
The scarcity of high-quality, annotated image data remains a bottleneck for the performance and generalisability of deep learning-based pose and depth estimation methods for microrobots~\cite{yang2024machine, shurrab2022self, plompen2020joint}.

\begin{figure*}[!t]
\centering
\includegraphics[width=1\hsize]{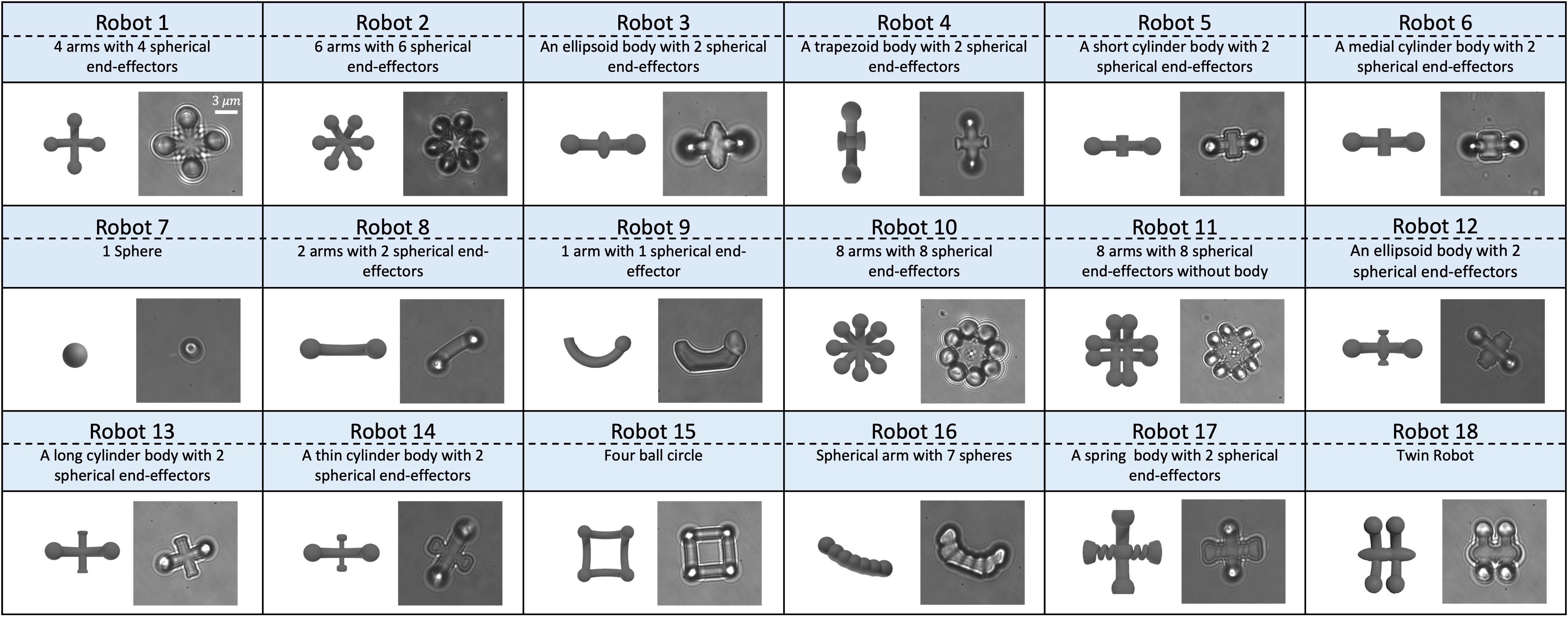}
\vspace{-0.4cm}
\caption{Overview of the 18 microrobot types included in the OTMR dataset. For each robot, the left image shows its CAD model, and the right image presents the corresponding experimental image captured at the focus plane under an optical microscope. Microrobots 1–6 (top row) are specifically designed for the pose classification task due to their varied and distinguishable orientations, while all 18 types are used for depth estimation tasks.
}
\vspace{-0.35cm}
\label{fig-cad}
\end{figure*}%

To fill this gap, we present the first publicly available dataset dedicated to microrobot perception under OT: the \underline{O}p\underline{T}ical \underline{M}icro\underline{R}obot dataset (OTMR).
OTMR is collected under microscopic imaging conditions and includes a broad spectrum of microrobot types, out-of-plane pose and depth variations, which capture the subtle and complex visual characteristics unique to micro-scale environments. 
In total, the dataset contains 232,881 images spanning 18 different microrobots (as shown in Fig.~\ref{fig-cad}) across 176 distinct poses.
OTMR is designed to support the development and evaluation of deep learning models for microrobot pose and depth estimation. To demonstrate its utility, we conduct a comprehensive set of experiments comparing multiple neural network architectures—including both standard and search-optimised models—on pose classification and depth regression tasks. 
Our analysis reveals how model depth and microrobot structural symmetry influence perception accuracy and highlights trends in performance across different dataset sizes, microrobot designs, and levels of visual complexity.
These findings offer insights into the design of effective microscale perception systems, paving the way for the development of closed-loop control systems for the autonomous operation of optical microrobots in microfluidic environments.


\section{Related Work}
In this section, we review both traditional computer vision-based approaches and deep learning–based methods for microrobot perception.

\begin{figure}[t!]
\centering
\includegraphics[width=0.9\hsize]{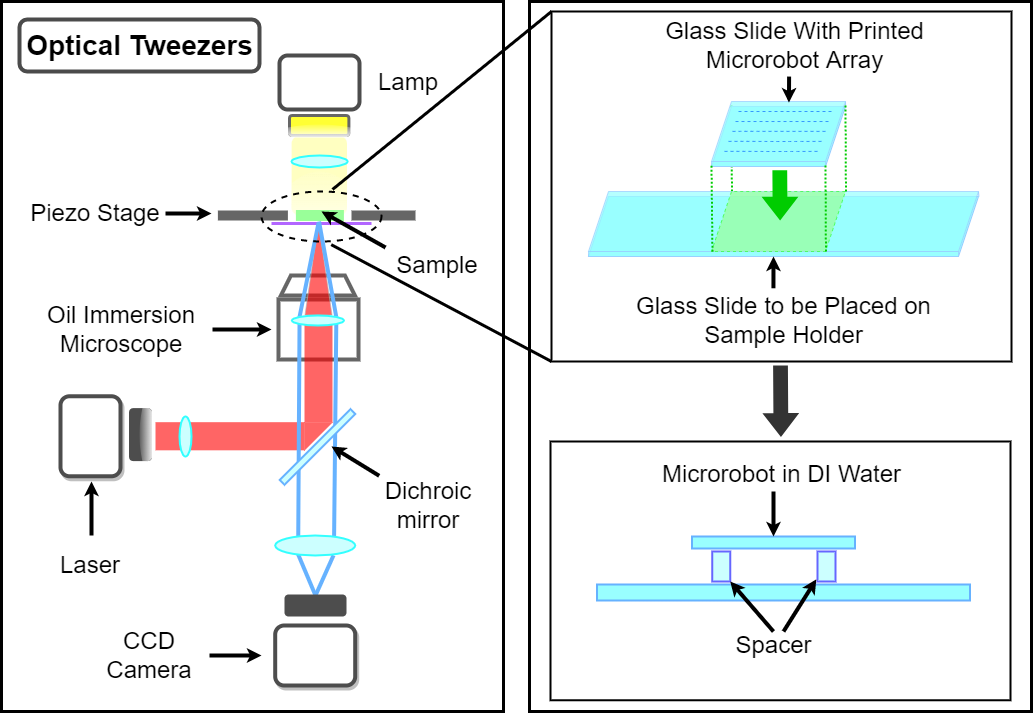}
\caption{Overview of the experimental platform for data collection. 
}
\label{fig:OT}
\end{figure}%

\begin{figure}[!t]
\centering
\includegraphics[width=1\hsize]{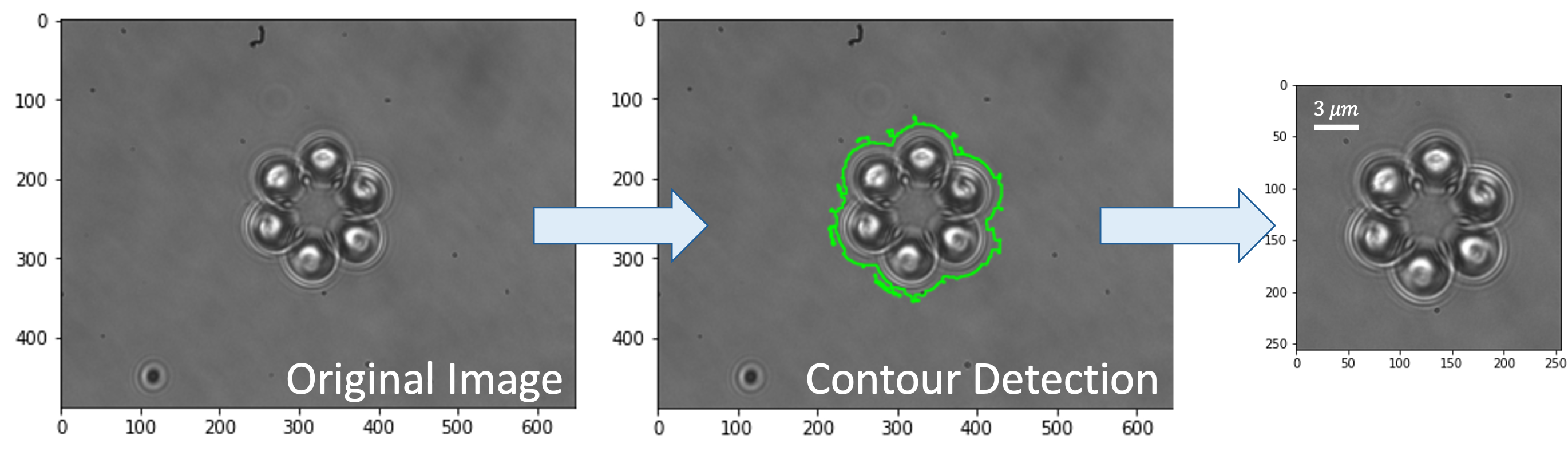}
\vspace{-0.4cm}
\caption{Preprocessing pipeline for microrobot image preprocessing. The original microscope image (left) is first processed through contour detection (middle) to identify the microrobot boundary. 
A $256 \times 256$ pixel region centred on the microrobot is then cropped (right) for subsequent analysis.}
\label{fig-preprocessing}
\vspace{-0.5cm}
\end{figure}%

\begin{figure*}[!t]
\centering
\includegraphics[width=1\hsize]{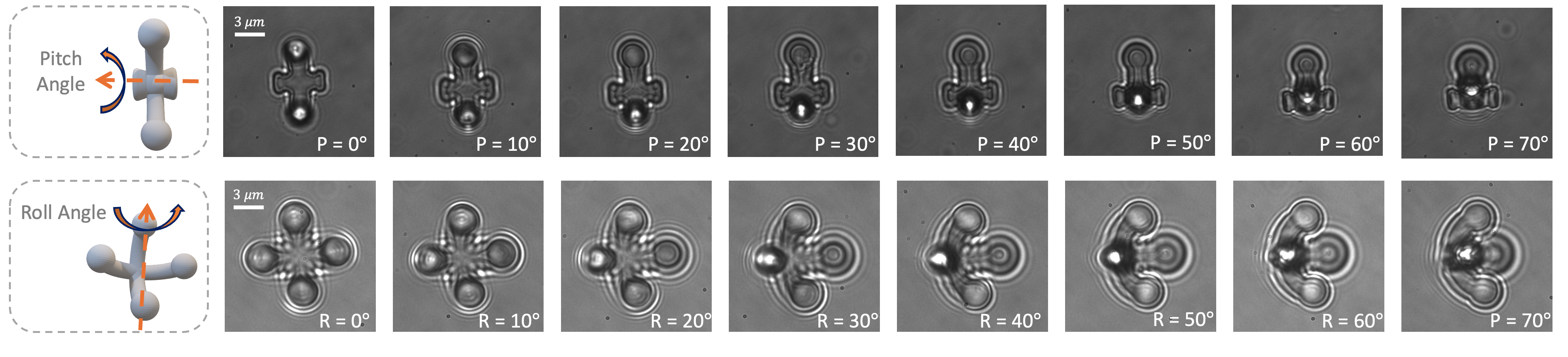}
\vspace{-0.7cm}
\caption{Illustration of pitch and roll angles in microrobot pose estimation. The top row shows variations in pitch angle (P), representing rotation around the horizontal axis, from $0^\circ$ to $70^\circ$. The bottom row shows variations in roll angle (R), representing rotation around the vertical axis, also from $0^\circ$ to $70^\circ$. }
\label{fig-pose_concept}
\vspace{-0.3cm}
\end{figure*}%

\begin{figure*}[!t]
\centering
\includegraphics[width=0.9\hsize]{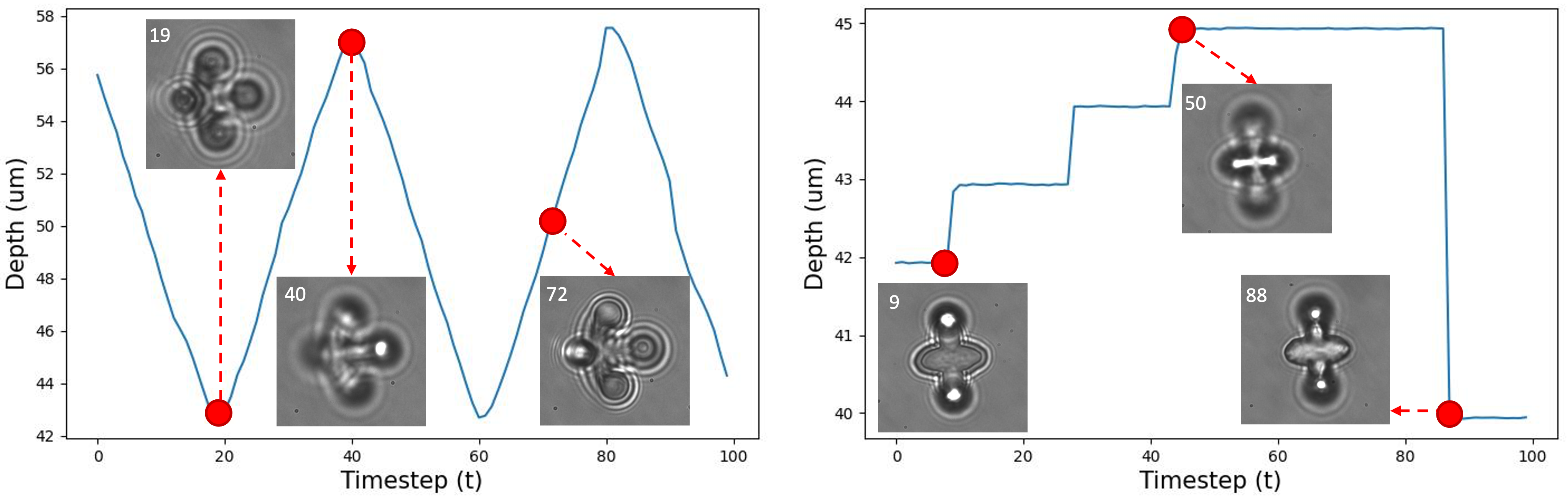}
\vspace{-0.4cm}
\caption{Visualisation of microrobot depth changes over time. The plots show depth (in $\mu$m) versus time steps for two different sequences. Sample frames at selected time steps are highlighted to illustrate the visual appearance of the microrobot at various depths, where depth is defined as the distance between the microrobot’s reference plane and the focal plane.}
\label{fig-depth_concept}
\vspace{-0.5cm}
\end{figure*}%

\subsection{Traditional Computer Vision-Based Methods}
Accurate depth and pose estimation are essential for closed-loop control of microrobots, enabling precise 3D manipulation of micro-scale targets such as cells and other biological objects~\cite{stavroulakis2016invited}. 
Traditional computer vision techniques have long been employed for micro-object perception tasks such as tracking, localisation, and pose estimation. 
These methods typically rely on handcrafted features, such as edges, contours, and geometric shapes, to detect and analyse microrobot structures under a microscope. 
Techniques like thresholding~\cite{seelaboyina2023different}, morphological operations~\cite{chudasama2015image}, and Hough or ellipse fitting~\cite{kanatani2016ellipse} have been used to identify and track components of microrobots, particularly in cases where the robots exhibit simple and symmetric designs. 
While these approaches can be effective in controlled environments with clear visual features and well-defined object boundaries, their performance often degrades in real-world microscale settings, where noise, optical diffraction, transparency, and low contrast are inherent characteristics of the imaging process~\cite{wu2024optical}.
Furthermore, traditional methods lack adaptability to new robot designs and require significant manual tuning, which limits their scalability and robustness in dynamic or complex experimental settings.
\subsection{Deep Learning Based Methods}
Supervised learning techniques have been increasingly adopted for achieving accurate and reliable perception of microscale objects~\cite{salehi2024advancements,rajasekaran2017accurate}. 
Grammatikopoulou et.al combined a convolutional neural network (CNN) with Long Short-Term Memory (LSTM) to estimate the 3D pose and depth of transparent microrobots observed under OT~\cite{grammatikopoulou2019three}. 
Zhang et.al integrated deep residual networks with Gaussian process regression to jointly estimate both spatial position and orientation~\cite{zhang2020data}. 
Despite these promising developments, the performance of machine learning models is heavily influenced by the quality and quantity of labelled training data. 
The lack of standardised, large-scale datasets limits reproducibility and hinders systematic benchmarking. 
To address this, we propose OTMR, the first publicly available dataset designed for microrobot perception under microscopy, and benchmark its performance using eight deep learning models applied to depth and pose estimation for microrobots.

\section{OTMR Dataset}
In this section, the microrobot fabrication, data collection and supported tasks are introduced. 
\subsection{Microrobot Fabrication}
The optical microrobots used in this study were fabricated using a Nanoscribe 3D printer (Nanoscribe GmbH, Germany) with IP-L 780 photoresist as the printing material. 
The fabrication process employed two-photon polymerisation (2PP)~\cite{kawata2001finer}. 
Microrobots were directly printed onto glass substrates and subsequently immersed in deionised (DI) water within a sealed spacer chamber for imaging and experimental use.

Eighteen distinct microrobot designs were fabricated for inclusion in the OTMR dataset. 
Their CAD models and corresponding focal plane images are shown in Fig.~\ref{fig-cad}. 
Microrobots 1–6 (top row) were primarily used for pose classification, featuring 176 unique out-of-plane poses generated by varying rotation angles from $0^\circ$ to $90^\circ$. 
All 18 types were used for depth estimation.
Microrobot geometry plays a crucial role in optical tweezers (OT)-based manipulation. 
In general, spherical and symmetrical structures are more readily trapped and stabilised under OT due to their uniform optical response and isotropic force distribution~\cite{gluckstad2011sculpting}. 
Geometries such as ellipsoids and cylinders have also been successfully demonstrated for optical trapping and translation~\cite{nieminen2001numerical}. 
However, achieving controlled out-of-plane rotation using conventional planar OT is challenging with simple sphere-like or symmetric geometries by changing the position of multiple laser spots~\cite{zhang2020distributed}.
To address this limitation, we systematically designed microrobots that enable out-of-plane rotation and depth control, allowing precise manipulation by modulating the position of laser foci within a planar OT system.

\subsection{Data Collection}
The data collection system is built around an OT platform, exemplified here by the setup from Elliot Scientific (UK), integrated with a nanopositioning stage (Mad City Labs Inc., USA). Microscopic images were captured using a high-speed CCD camera (Basler AG, Germany) mounted on a Nikon Ti microscope with a 100× oil immersion objective. Each image frame has a resolution of 678×488 pixels. 
Although the system described is based on a specific OT configuration, the data collection methodology is generalisable and can be implemented in other micromanipulation systems, such as optical tweezers with integrated microscopes or SEM-based platforms equipped with micro-grippers for physical manipulation.
A schematic of the system is shown in Fig.~\ref{fig:OT}.

For optical microscope-based imaging, microrobots with fixed poses were translated along the $z$-axis using the piezoelectric stage to simulate varying focal depths. This approach captures the natural defocus and blur encountered at different distances from the focal plane, generating depth-varying image sequences for each pose. The motion range was limited to remain within the vertical extent of the microrobot to ensure reliable depth variation while preserving visibility.

Following collection, raw microrobot images were first converted to grayscale and denoised using a pixel-wise adaptive low-pass Wiener filter~\cite{lim1990two}. 
Binarisation was then applied using an automatically selected threshold, followed by edge detection using the Canny algorithm~\cite{canny1986computational}. 
The centroid of each microrobot was identified, and a $256 \times 256$ pixel region of interest (ROI) centred on the microrobot was cropped for subsequent analysis. 
The complete preprocessing pipeline is depicted in Fig.~\ref{fig-preprocessing}.

\subsection{Supported Tasks}
\subsubsection{Pose Classification}
One of the primary tasks supported by the OTMR dataset is pose classification, which aims to determine the out-of-plane orientation of microrobots. 
As shown in Fig.~\ref{fig-pose_concept}, each microrobot can rotate along two axes: pitch and roll. 
We fabricated 176 unique poses by varying pitch and roll angles from 0° to 90°, capturing their corresponding microscope images under an optical tweezer system. 
These variations result in diverse appearance changes due to optical diffraction and focus distortion, posing a significant challenge for traditional image-based classification methods.

\subsubsection{Depth Regression}
In addition to pose estimation, the OTMR dataset also enables the task of depth regression, which involves predicting the microrobot’s vertical position relative to the imaging plane. 
As illustrated in Fig.~\ref{fig-depth_concept}, depth changes are induced by controlling a piezoelectric stage that moves the microrobot along the z-axis. 
We record the true depth values at each timestep while simultaneously capturing microscopic images. 
OTMR includes both continuous and stepwise depth trajectories, providing data that reflects realistic experimental conditions and dynamic visual appearances. 
This task is critical for enabling closed-loop feedback control in 3D micromanipulation, and OTMR serves as a comprehensive benchmark for evaluating depth regression performance under microscopic constraints.

\section{Experiments}
\begin{table}[!t]
\centering
\caption{Neural architecture search space used in Optuna for optimising CNN models. }
\resizebox{\linewidth}{!}{
\begin{tabular}{c|c|c|c}
\hline \hline 
Hyperparameter & Type & Range & Description\\
\hline
num conv& Integer& 1 $-$ 9& No.convolutional blocks\\ 
num filters& Integer& 16 $-$ 256& No.channels in each convolutional layer\\ 
fc dim& Integer& 64 $-$ 1024& No.neurons in the fully connected layer\\ 
dropout rate& Float& 0.0 $-$ 0.5& Dropout rate applied before classification\\ 
use BN& Categorical& True, False& Whether to apply Batch Normalization\\ 
learning rate& Float& 1e-5 $-$ 1e-2&Learning rate for the Adam optimizer \\ 
batch size& Categorical& 8, 16, 32, 64& Batch size used in training\\ 
\hline \hline 
\end{tabular}}
\label{table-optuna}
Note: These hyperparameters are tuned using Bayesian optimisation to maximise pose classification accuracy or minimise depth regression error.
\vspace{-0.5cm}
\end{table}
\subsection{Implementation Details}
All experiments were implemented using PyTorch 1.10 and Python 3.9, and executed on a system equipped with a single NVIDIA A100 GPU (80 GB memory). 
The CUDA version was 12.0, and all throughputs were measured during inference on an A100 GPU with a batch size of 128 and 32-bit floating point precision.
To address the unique challenges of microscale perception, such as visual contrast, limited features, and structural symmetry, we benchmark a representation set of deep learning models for pose classification and depth regression.
These include convolutional-based architectures (such as CNN, VGG16~\cite{simonyan2014very}, ResNet18, and ResNet50~\cite{he2016deep}) for their strong spatial feature extraction capabilities, and lightweight models (EfficientNet~\cite{efficientnet} and MobileNetV2~\cite{mobilenet}) for real-time, resource-efficient applications.
We also evaluate a transformer-based model, ViT~\cite{dosovitskiy2020image}, which captures global dependencies and performs well in visually ambiguous scenarios. 
Additionally, we apply a NAS method to optimise a CNN model architecture for microscale tasks.

The NAS method uses Bayesian optimisation via Optuna~\cite{akiba2019optuna}, with a defined search space (detailed in Table~\ref{table-optuna}).
We perform separate architecture searches for the pose classification and depth regression tasks, each running for 150 epochs. 
For pose classification, the search objective is to maximise the average classification accuracy of pitch and roll angles. For depth regression, the objective is to minimise the mean squared error (MSE) between predicted and ground truth depth values.

CNN models and NAS-optimised architectures are trained from scratch for 20 epochs. For all other models, we fine-tune pre-trained weights available from the \textit{torchvision} library~\cite{torchvision2016}, also for 20 epochs, using a fixed learning rate of 1e-3.
To ensure fair and robust evaluation, we adopt a five-fold cross-validation strategy across all benchmark experiments. 
The dataset is evenly partitioned into five subsets; in each fold, one subset is reserved for testing, while the remaining four are split into training and validation sets using a fixed per-class sampling strategy. 
Final results are reported as the average performance across all five folds.

\begin{table*}[!t]
\centering
\caption{Pose classification five-fold cross-validation results for Robot 1 and Robot 3 across all models.}
\resizebox{\linewidth}{!}{
\begin{tabular}{c|cc:cc:cc:cc|cc:cc:cc:cc}
\hline \hline 
\multirow{3}{*}{Method} & \multicolumn{8}{|c}{Robot 1} & \multicolumn{8}{|c}{Robot 3}\\ 
\cline{2-17}
& \multicolumn{2}{c:}{Accuracy($\uparrow$)} & \multicolumn{2}{c:}{Precision($\uparrow$)} & \multicolumn{2}{c:}{Recall($\uparrow$)} & \multicolumn{2}{c|}{F1 Score($\uparrow$)} & \multicolumn{2}{c:}{Accuracy($\uparrow$)} & \multicolumn{2}{c:}{Precision($\uparrow$)} & \multicolumn{2}{c:}{Recall($\uparrow$)} & \multicolumn{2}{c}{F1 Score($\uparrow$)} \\
\cline{2-17}
& Pitch & Roll & Pitch & Roll  & Pitch & Roll& Pitch & Roll& Pitch & Roll & Pitch & Roll  & Pitch & Roll& Pitch & Roll \\ \hline
CNN & 0.979 & 0.981 & 0.979 & 0.975 & 0.972 & 0.969 & 0.975 & 0.971
& 0.845 & 0.941 & 0.781 & 0.915 & 0.776 & 0.915 & 0.761 & 0.913 \\ 
VGG16 & \underline{0.996} & 0.989 & \underline{0.997} & 0.991 & \underline{0.997} & 0.992 & \underline{0.997} & 0.991
& \underline{0.955} & \underline{0.970} & \underline{0.922} & \textbf{0.961} & \underline{0.933} & \underline{0.958} & \underline{0.922} & \underline{0.958} \\ 
ResNet18 & 0.876 & 0.836 & 0.967 & 0.876 & 0.865 & 0.828 & 0.860 &  0.815 
& 0.904 & 0.932 & 0.913 & 0.918 & 0.892 & 0.907 & 0.885 & 0.903 \\ 
ResNet50 & 0.989 & 0.991 & 0.989 & 0.991 & 0.989 & 0.989 & 0.988 &  0.990 
& 0.819 & 0.863 & 0.807 & 0.867 & 0.742 & 0.830 &0.725 & 0.825 \\ 
EfficientNet & 0.988 & 0.986 & 0.986 &0.989 & 0.988 & 0.987 & 0.986 & 0.988  
& 0.930 & 0.966 & 0.893 & 0.950 & 0.894 & 0.954 & 0.881 & 0.951 \\ 
MobileNetV2 & 0.985 & 0.991 & 0.987 & 0.992 & 0.984 & 0.991 & 0.985 & 0.991  
& 0.944 & 0.961 & 0.920 & 0.948 & 0.928 & 0.946 &0.918  & 0.943 \\ 
ViT & \textbf{0.999} & \textbf{0.998} & \textbf{0.999} & \textbf{0.998} & \textbf{0.999} & \textbf{0.998} & \textbf{0.999} & \textbf{0.998 } 
& \textbf{0.965} & \textbf{0.971} & \textbf{0.943} & \underline{0.960} & \textbf{0.950} & \textbf{0.959}& \textbf{0.944} & \textbf{0.959} \\ 
NAS & 0.981& \underline{0.993}& 0.981& \underline{0.993}& 0.978& \underline{0.994}& 0.979& \underline{0.993}& 0.922& 0.959& 0.907& 0.943& 0.879& 0.940& 0.888&  0.941\\ 
\hline \hline 
\end{tabular}}\\
Note: The best results in each column are highlighted in \textbf{bold} and the second‐best are \underline{underlined}.
$\downarrow$ indicates that the lower the better and $\uparrow$ the opposite. 
\label{table-pose_result}
\vspace{-0.35cm}
\end{table*}

\begin{table}[!t]
\centering
\caption{Comparison of model size (MB), inference cost (GFLOPs), and throughput (images/sec) for pose classification across benchmarked methods.}
\begin{tabular}{c|c:c:c}
\hline \hline 
Method & Params (MB) & GFLOPs &  Throughput (img/s) \\
\hline
CNN         &25.79   & 0.53  &  15509.49 \\ 
VGG16       & 134.33 & 15.47 &  2130.84  \\
ResNet18    & 11.18  & 1.82  &  10725.36 \\
ResNet50    & 23.54  & 4.13  &  3082.20  \\
EfficientNet& 4.03   & 0.41  &  4971.11  \\
MobileNetV2 & 2.24   & 0.33  &  6987.17  \\
ViT         & 86.25  & 16.86 &  1385.27  \\
NAS         & 33.66  & 3.11  &  3007.85   \\
\hline \hline 
\end{tabular}
\label{table-model_size}
Note: Throughput refers to the number of images processed per second.
\vspace{-0.5cm}
\end{table}

\begin{table*}[!t]
\centering 
\caption{Depth regression results for robots 8-18 across all models.}
\resizebox{\linewidth}{!}{
\begin{tabular}{c|cc:cc:cc:cc:cc:cc}
\hline \hline 
\multirow{2}{*}{Method} & \multicolumn{2}{|c:}{Robot 8} & \multicolumn{2}{c:}{Robot 10} & \multicolumn{2}{c:}{Robot 12} & \multicolumn{2}{c:}{Robot 14}& \multicolumn{2}{c:}{Robot 16}& \multicolumn{2}{c}{Robot 18} \\
\cline{2-13}
 & MSE($\downarrow$) & $R^2$($\uparrow$) & MSE($\downarrow$) & $R^2$($\uparrow$)  & MSE($\downarrow$) & $R^2$($\uparrow$) & MSE($\downarrow$) & $R^2$($\uparrow$) & MSE($\downarrow$) & $R^2$($\uparrow$) & MSE($\downarrow$) & $R^2$($\uparrow$) \\ \hline 
 CNN          & 0.052& 0.996& 0.479& 0.965& 0.104& 0.992& 0.520& 0.978& 0.219& 0.983& 0.852& 0.940\\ 
 VGG16        & 0.113& 0.994& 0.504& 0.963& 0.191& 0.985& 0.371& 0.984& 0.364& 0.972& 0.311& 0.978\\
 ResNet18     & 0.053& 0.996& 0.082& \underline{0.994}& \underline{0.081}& \underline{0.993}& 0.299& 0.987& 0.063& 0.995& 0.059& 0.995\\
 ResNet50     & \textbf{0.043}& \textbf{0.998}& \underline{0.075}& \underline{0.994}& \textbf{0.074}& \textbf{0.994}& 0.307& 0.987& \underline{0.034}& \underline{0.997}& 0.054& \underline{0.996}\\
 EfficientNet & 0.048& 0.996& \textbf{0.068}& \textbf{0.995}& 0.101& 0.992& \underline{0.238}& \textbf{0.990}& 0.084& 0.993& 0.078& 0.994\\
 MobileNetV2  & 0.055& 0.996& 0.093& 0.993& 0.098& 0.992& 0.253& \underline{0.989}& 0.064& 0.995& 0.186& 0.986\\
 ViT          & 0.054& 0.996& 0.076& \underline{0.994}& 0.096& 0.992& \textbf{0.232}& \textbf{0.990}& 0.048& 0.996& \textbf{0.011}& \textbf{0.999}\\
 NAS          & \underline{0.046}& \underline{0.997}& 0.118& 0.992& 0.087& \underline{0.993}& 0.258& \underline{0.989}& \textbf{0.017}& \textbf{0.998}& \underline{0.051} &\underline{0.996}\\
\hline \hline 
\end{tabular}}\\
\vspace{-0.2cm}
\begin{center}
\begin{minipage}{0.88\textwidth}
  \hspace{1.05cm} 
  \includegraphics[width=\linewidth]{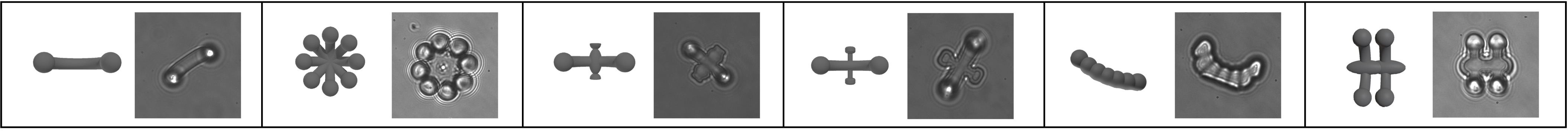}
\end{minipage}
\end{center}
\vspace{-0.2cm}
Note: The MSE values are in $\mu m$. The best results in each column are highlighted in \textbf{bold} and the second‐best are \underline{underlined}.
$\downarrow$ indicates that the lower the better and $\uparrow$ the opposite.
Figures on the last row are the CAD model and the corresponding experimental image of each robot.
\label{table-depth_result}
\vspace{-0.2cm}
\end{table*}

\subsection{Evaluation Metrics}
We employ task-specific evaluation metrics to assess model performance on pose classification and depth regression.
For the pose classification task, we use four standard metrics:
1) Accuracy: the proportion of correctly predicted pose labels over the total number of predictions.
2) Precision: the proportion of correctly predicted positive instances among all predicted positives, reflecting prediction exactness.
3) Recall: the proportion of correctly predicted positive instances among all actual positives, indicating the model’s sensitivity.
4) F1 Score: the harmonic mean of precision and recall, providing a balanced measure of model performance.

For the depth regression task, we use the following metrics:
1) MSE: the average squared difference between predicted and ground truth depth values, which penalises larger errors more heavily.
2) Coefficient of Determination ($R^2$): which measures how well the predicted values explain the variance in the ground truth data. An $R^2$ value closer to 1 indicates better fit.

In addition to predictive performance, we report the model size in terms of the number of parameters (in megabytes, MB), the computational complexity in floating point operations per second (FLOPs), measured in gigaflops (GFLOPs) and real-time processing capability measured as throughput (the number of images that can be processed per second). 
These metrics help evaluate the trade-off between accuracy and computational efficiency.

\subsection{Benchmarking} 
Table~\ref{table-pose_result} presents the five-fold cross-validation results for pose classification using two different microrobot types, with models trained on an equal number of images per pose. 
The results indicate that microrobots with more complex structures, such as Robot 3, which incorporates two distinct types of spherical components, pose greater challenges for pose estimation compared to simpler designs like Robot 1, which consists of four identical spheres. 
For instance, the best pitch and roll prediction accuracies for Robot 3 are 3.4\% and 2.7\% lower, respectively, than those for Robot 1.
Among all evaluated architectures, the ViT consistently outperforms the others across different microrobot types. 
While ViT, like many convolution models such as ResNet, benefits from pretraining on the large-scale ImageNet dataset~\cite{deng2009imagenet}, its architectural differences contribute more to its performance in microscale tasks.
Specifically, ViT's patch-based tokenisation and self-attention mechanism allow it to model both local details and long-range spatial dependencies more effectively. These capabilities are particularly advantageous in micro-robotic images, where visual features are subtle, spatially distributed, and often lack strong texture or contrast.
Table~\ref{table-model_size} compares the computational characteristics of various models for the pose classification task. 
Although the ViT exhibits the highest GFLOPs among all models, it is still capable of processing over 1,300 images per second, demonstrating strong real-time performance despite its computational intensity.

Table~\ref{table-depth_result} summarises the depth regression results for six different microrobot types. Similar to pose classification, robots with complex and asymmetric geometries (e.g., Robot 14) are significantly more difficult to regress accurately. The lowest MSE obtained on Robot 14 is approximately six times higher than that of a simpler design like Robot 8. Furthermore, for a given robot, deeper architectures (e.g., ResNet50) tend to outperform shallower ones (e.g., ResNet18).

The results of NAS are shown in Table~\ref{table-optuna-value}. The NAS process was applied to CNN-based architectures and trained from scratch. As reported in Tables~\ref{table-pose_result} and~\ref{table-depth_result}, the NAS-optimised models consistently outperform the baseline CNN in all evaluated cases. Notably, the NAS model achieved the best depth regression performance on Robot 16, one of the complex designs, demonstrating the effectiveness of architecture search in tailoring models to task difficulty.
Moreover, the architectures discovered by NAS further reflect the relative difficulty of the tasks. The optimal model for depth regression includes two additional convolutional layers and a larger fully connected layer compared to the model optimised for pose classification, aligning with the inherently more complex nature of continuous-value prediction in regression tasks.

\begin{table}[!t]
\centering
\caption{Optimal architectures found by NAS for pose classification and depth regression.}
\begin{tabular}{c:c|c:c}
\hline \hline 
\multicolumn{2}{c}{Pose Task} & \multicolumn{2}{|c}{Depth Task}\\ \hline
Hyperparameter & Value & Hyperparameter & Value\\
\hline
num conv& 3& num conv&5\\ 
num filters& 142& num filters&100\\ 
fc dim& 299& fc dim&493\\  
dropout rate& 0.2& dropout rate&0.31\\ 
use BN&True  & use BN&True\\  
learning rate&1.4e-4 & learning rate&4.8e-4\\ 
batch size& 8& batch size&32\\ 
\hline
Params (MB) &  33.66 & Params (MB)& 9.71\\
GFLOPs &  3.11 & GFLOPs &5.85 \\
\hline \hline 
\end{tabular}
\label{table-optuna-value}
\end{table}

\subsection{Transfer Learning Among Different Robots}
To evaluate the generalisation ability of deep learning models, we conducted a transfer learning experiment using the best-performing model, ViT, trained on data from Robot Type 3 for the pose classification task. The trained model was directly tested on Robot Types 1, 4, and 5 without further fine-tuning. The evaluation metric is the average classification accuracy of pitch and roll angles.

\begin{figure}[!t]
\centering
\includegraphics[width=1\hsize]{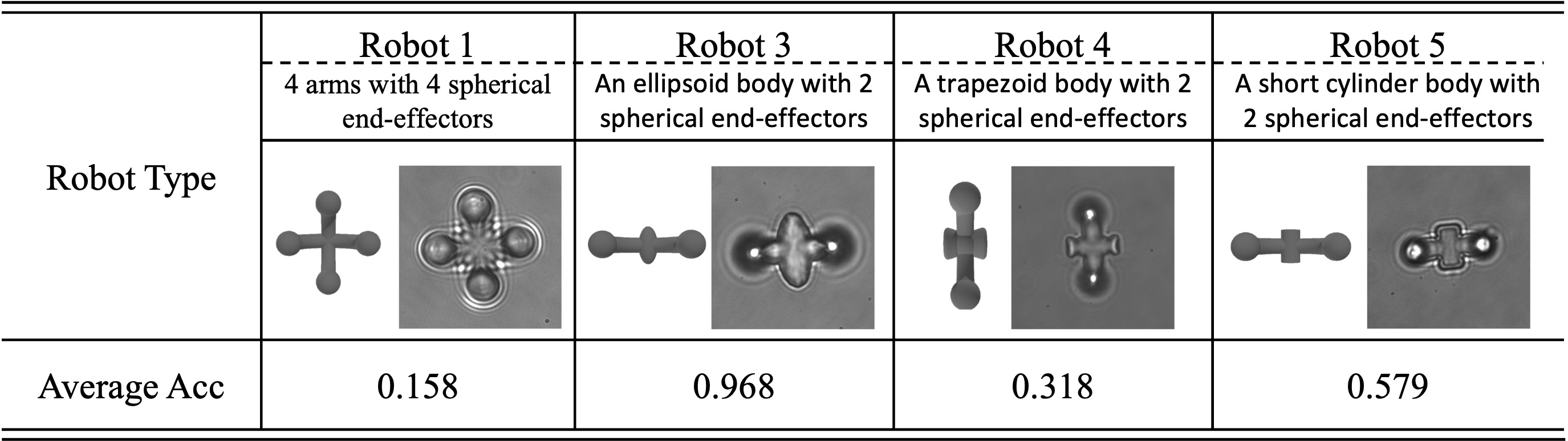}
\vspace{-0.3cm}
\caption{Transfer learning results of the ViT model trained on Robot Type 3 and tested on different robot types without fine-tuning. The evaluation metric is the average classification accuracy of pitch and roll angles.}
\label{fig-transfer}
\vspace{-0.2cm}
\end{figure}

As shown in Fig.~\ref{fig-transfer}, the model achieves the highest accuracy on Robot 3, the training target, as expected. 
Robots 4 and 5 exhibit higher classification accuracy than Robot 1, likely due to structural similarity to Robot 3. All three robots (3, 4, and 5) share a common feature: they are composed of two distinct types of spherical components along the arms. 
In contrast, Robot 1 consists of four identical spheres, which differ significantly in geometry and visual features.
Furthermore, the spatial orientation of the robot also affects transfer performance. 
Robot 5 shares the same horizontal configuration as Robot 3, leading to better generalisation and higher accuracy. 
In contrast, Robot 4 is vertically oriented, which introduces a distribution shift in visual appearance and results in reduced classification performance compared to Robot 5.

\subsection{Model Interpretability}

\begin{figure}[!t]
\centering
\includegraphics[width=0.95\hsize]{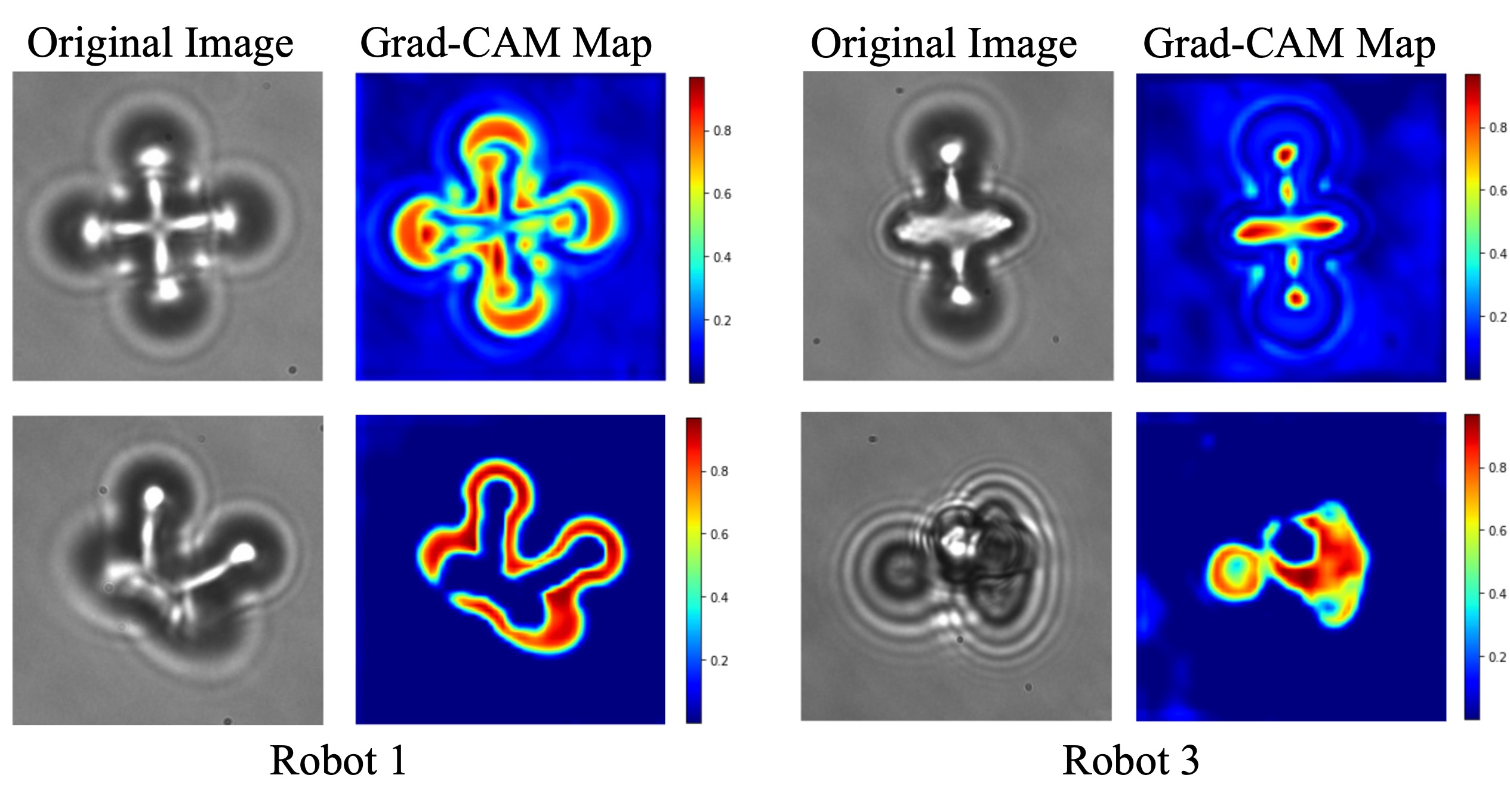}
\vspace{-0.3cm}
\caption{Grad-CAM visualisations for pose classification on Robot Types 1 and 3 using a CNN model. Each pair shows the original microscope image (left) and its corresponding Grad-CAM heatmap (right). The red regions indicate high-importance areas that the model relies on most for its predictions, while blue regions indicate areas of low importance that are largely ignored. }
\label{fig-grad_cam}
\vspace{-0.5cm}
\end{figure}%

To gain insights into which regions of an input microrobot image influence the model’s predictions, we employ Gradient-weighted Class Activation Mapping (Grad-CAM)~\cite{selvaraju2017grad} to visualise the spatial attention of the CNN before the last fully connected layer during pose classification. 
As illustrated in Fig.~\ref{fig-grad_cam}, the Grad-CAM heatmaps highlight the areas that contribute most to the model’s decision-making process.

The visualisations for Robot Types 1 and 3 reveal that the model consistently attends to the microrobot structure itself, particularly the arms and spherical components, when determining the pose. 
This confirms that CNN has successfully learned to focus on the relevant features of the microrobot rather than background noise, providing interpretability and confidence in the model’s classification behaviour.

\subsection{Influence of Data Size}
\begin{figure}[!t]
\centering
\includegraphics[width=1\hsize]{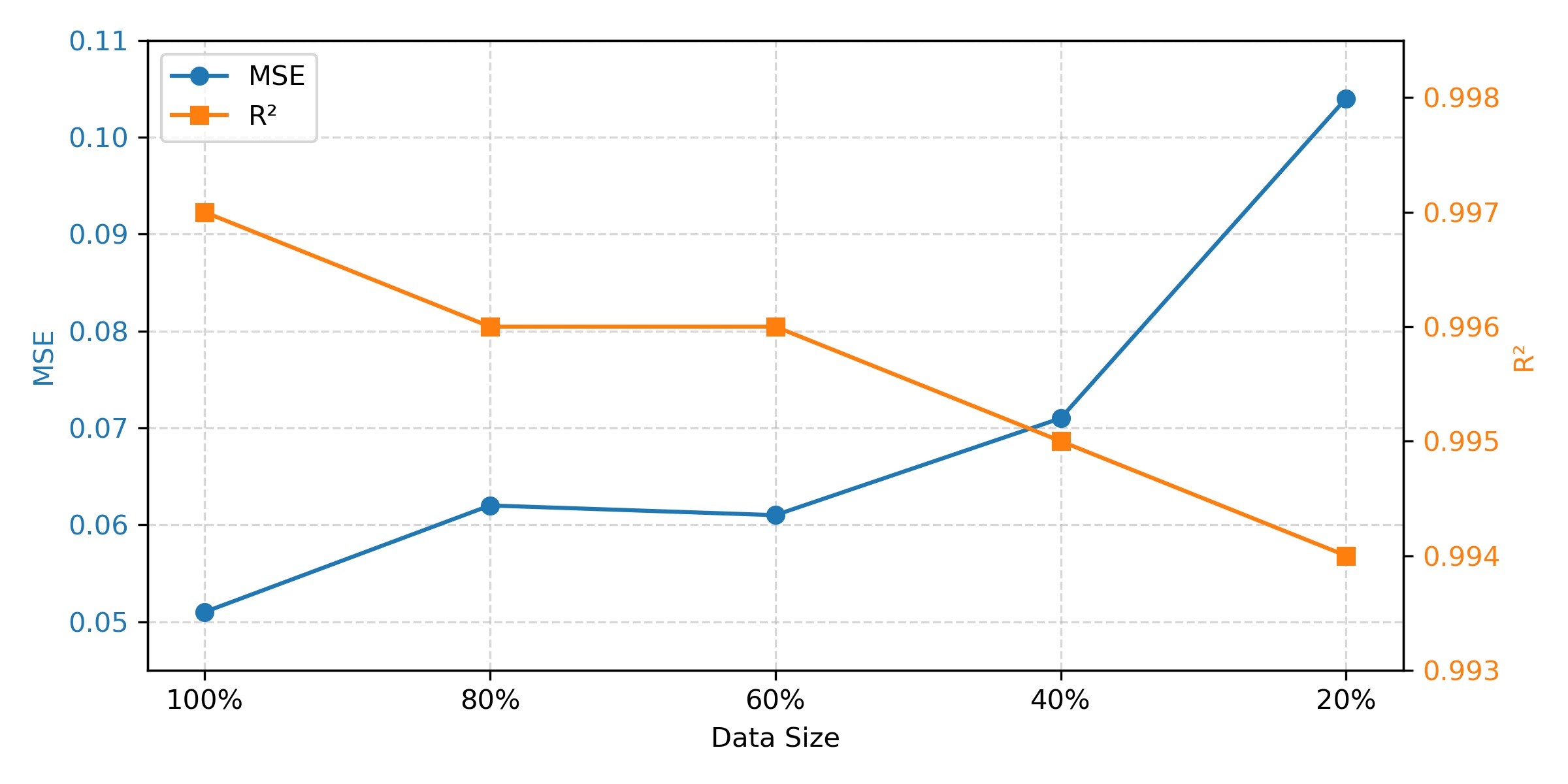}
\vspace{-0.5cm}
\caption{Impact of training data size on depth regression performance using ResNet50 for Robot Type 8.}
\label{fig-data_size}
\vspace{-0.5cm}
\end{figure}%

To evaluate the impact of dataset size on depth regression performance, we conduct experiments using Robot Type 8 and the best-performing model identified in Table~\ref{table-depth_result}, ResNet50, trained for 10 epochs. 
The complete dataset for this robot consists of 5,600 images. 
We train and test the model using varying proportions of the data: 100\%, 80\%, 60\%, 40\%, and 20\%, while maintaining a fixed train/validation/test split of 8:1:1 in each case.
As shown in Fig.~\ref{fig-data_size}, increasing the amount of training data consistently reduces the MSE and improves the $R^2$ score, indicating better regression accuracy and stronger predictive reliability. 
These results emphasise the importance of large-scale data availability for training deep learning models in micro-scale environments and demonstrate the value of the OTMR dataset in enabling robust, data-driven depth estimation.

\section{Discussion}
Our experiments use standard metrics such as accuracy, precision, recall, and F1 score for pose classification, and MSE and $R^2$ for depth regression. However, these metrics may not fully reflect the performance needs in microscale tasks. 
For example, small prediction errors near class boundaries or slight depth deviations may not significantly impact real-world micromanipulation but can still reduce metric scores.
Grad-CAM visualisations help interpret model behaviour by highlighting key regions used for prediction. 
These insights can inform future model design, such as incorporating attention mechanisms, or guide targeted data augmentation to improve robustness, especially in challenging visual conditions.

\section{Future Research Directions}
The OTMR dataset enables several promising directions for advancing microscale perception in biomedical applications. A key focus is simulation-to-reality (sim-to-real) transfer learning, which reduces the need for labour-intensive data collection by training models in simulation and adapting them to real-world conditions. To support this, we provide open-source CAD models that can be used to construct digital twins~\cite{tao2022digital} and generate synthetic microrobot images with realistic variations in pose, depth, illumination, and noise.
Generative AI methods, such as GANs~\cite{goodfellow2020generative} and diffusion models~\cite{croitoru2023diffusion}, can further enhance realism in synthetic data, improving sim-to-real generalisation. 
These approaches also support closed-loop and autonomous control by enabling real-time feedback using models trained primarily in simulated environments.
Another direction is cross-robot generalisation, exploring how structural similarity impacts transfer learning performance. For example, one can investigate how few-shot adaptation from a source robot enables accurate perception on new designs, ultimately reducing the retraining burden for deploying autonomous microrobots at scale.
Instead of relying on macro-scale pretraining (e.g., ImageNet~\cite{deng2009imagenet}), OTMR offers a domain-specific pretraining resource tailored to microscale visual tasks. Models trained on OTMR can serve as backbones for downstream applications such as micro-object tracking, 3D reconstruction, and control in dynamic environments.

All open-source resources—including the dataset, models, and benchmarking codes—are available at the project website: \href{https://lannwei.github.io/Optical-Microrobot-Database/}{Optical Microrobot Database}, and will be continuously updated to support future research.

\section{Conclusion}
In this work, we introduced OTMR, the first publicly available dataset designed specifically for microrobot perception under optical tweezers. 
OTMR captures the unique visual characteristics of microscale environments through 232,881 images across 18 microrobot types and 176 poses. 
We conducted comprehensive benchmarks across multiple deep-learning models for both pose classification and depth regression,  
Results show that ViT achieved the highest pose classification accuracy, exceeding 99\%, while depth regression tasks benefited significantly from deeper architectures. 
Moreover, increasing the size of the training dataset consistently improved model performance in both tasks.
These findings highlight the inherent challenges of visual perception at the microscale and emphasise the importance of domain-specific datasets like OTMR in advancing robust, data-driven solutions for perception and closed-loop control in micro-scale systems.


\section*{Acknowledgement}
The authors would like to thank Mr. Wen Fan for insightful discussions and figure advice.
Lan Wei acknowledges support from a PhD studentship jointly sponsored by the China Scholarship Council, and the Department of Bioengineering, Imperial College London.

\bibliographystyle{IEEEtran}
\bibliography{main}

\end{document}